\def\year{2021}\relax
\documentclass[letterpaper]{article} 
\usepackage{aaai21}  
\usepackage{times}  
\usepackage{helvet} 
\usepackage{courier}  
\usepackage[hyphens]{url}  
\usepackage{graphicx} 
\usepackage{natbib}  
\usepackage{caption} 
\usepackage{booktabs}       
\usepackage{amsmath,amssymb} 
\usepackage{amsfonts}       
\usepackage{nicefrac}       
\usepackage{microtype}      
\usepackage{comment}
\usepackage{xfrac}
\usepackage{esvect}
\usepackage{epsfig}
\usepackage{xspace}
\usepackage{nccmath}
\usepackage{adjustbox}		
\usepackage{calc}
\usepackage{tabularx}
\usepackage{makecell}
\usepackage{multirow}
\usepackage[hang,flushmargin]{footmisc}
\usepackage[neverdecrease]{paralist}
\usepackage{subcaption}
\usepackage{algorithm}
\usepackage{algorithmic}
\usepackage[table]{xcolor}
\usepackage{colortbl}
\usepackage{pifont}
\usepackage{animate}

\usepackage{hyperref}

\newcommand{\calF}{{\mathcal{F}}}

\newcommand{\calL}{{\mathcal{L}}}

\newcommand{\be}{\begin{eqnarray}}
\newcommand{\ee}{\end{eqnarray}}
\newcommand{\bee}{\begin{eqnarray*}}
\newcommand{\eee}{\end{eqnarray*}}

\newcommand{\matrixb}{\left[ \begin{array}}
\newcommand{\matrixe}{\end{array} \right]}

\definecolor{CuGray}{gray}{0.9}
\newcolumntype{g}{>{\columncolor{CuGray}}c}

\newcommand{\cmark}{\ding{51}}%
\newcommand{\xmark}{\ding{55}}%

\makeatletter
\DeclareRobustCommand\onedot{\futurelet\@let@token\@onedot}
\def\@onedot{\ifx\@let@token.\else.\null\fi\xspace}

\newcommand{\dashrule}[1][black]{%
  \color{#1}\rule[\dimexpr.5ex-.2pt]{4pt}{.4pt}\xleaders\hbox{\rule{4pt}{0pt}\rule[\dimexpr.5ex-.2pt]{4pt}{.4pt}}\hfill\kern0pt%
}

\newcommand*{\Scale}[2][4]{\scalebox{#1}{$#2$}}%

\newcommand{\figref}[1]{Fig.~\ref{#1}}
\newcommand{\tabref}[1]{Table~\ref{#1}}
\newcommand{\eqnref}[1]{Eq.~(\ref{#1})}

\def\eg{\emph{e.g}\onedot}

\def\etal{\emph{et al}\onedot}

\title{Learning Monocular Depth in Dynamic Scenes via\linebreak Instance-Aware Projection Consistency}
\author{
    Seokju Lee$^{1}$ \hspace{6mm}
    Sunghoon Im$^{2}$ \hspace{6mm}
    Stephen Lin$^{3}$ \hspace{6mm}
    In So Kweon$^{1}$ \\
}
\affiliations{
    $^{1}$KAIST \hspace{6mm}
    $^{2}$DGIST \hspace{6mm} 
    $^{3}$Microsoft Research Asia \\
    \texttt{seokju91@gmail.com} \\
}

\begin{document}

\maketitle

\begin{abstract}
We present an end-to-end joint training framework that explicitly models 6-DoF motion of multiple dynamic objects, ego-motion and depth in a monocular camera setup without supervision. Our technical contributions are three-fold. First, we highlight the fundamental difference between inverse and forward projection while modeling the individual motion of each rigid object, and propose a geometrically correct projection pipeline using a neural forward projection module. Second, we design a unified instance-aware photometric and geometric consistency loss that holistically imposes self-supervisory signals for every background and object region. Lastly, we introduce a general-purpose auto-annotation scheme using any off-the-shelf instance segmentation and optical flow models to produce video instance segmentation maps that will be utilized as input to our training pipeline. These proposed elements are validated in a detailed ablation study. Through extensive experiments conducted on the KITTI and Cityscapes dataset, our framework is shown to outperform the state-of-the-art depth and motion estimation methods. Our code, dataset, and models are available at \url{https://github.com/SeokjuLee/Insta-DM}.
\end{abstract}


\vspace{-0mm}
\section{Introduction}
\vspace{-0mm}

Knowledge of the 3D environment structure and the motion of dynamic objects is essential for autonomous navigation~\cite{shashua2004pedestrian,geiger20143d}.
The 3D structure is valuable because it implicitly models the relative position of the agent, and it is also utilized to improve the performance of high-level scene understanding tasks such as detection and segmentation~\cite{lee2015rich,lee2017vpgnet,yang2018lego,shin2019roarnet,behley2019iccv,lee2019visuomotor}.
Besides scene structure, the 3D motion of the agent and traffic participants such as pedestrians and vehicles is also required for safe driving.
The relative direction and speed between them are taken as the primary inputs for determining the next direction of travel.

\begin{figure}[t]
    \centering
    \begin{subfigure}[b]{0.999\linewidth}
        \centering
        \animategraphics[autoplay,loop,width=0.495\linewidth]{21}{./figure/inv_bremen_127/}{0000}{0109}
        \hfill
        \animategraphics[autoplay,loop,width=0.495\linewidth]{21}{./figure/fwd_bremen_127/}{0000}{0109}
        \caption{Inverse projection~\cite{casser2019depth} and forward projection.}
        \label{teaser_a}
    \end{subfigure}
    \vskip 0.7\baselineskip
    \begin{subfigure}[b]{0.999\linewidth}
        \centering
        \animategraphics[autoplay,loop,width=0.495\linewidth]{21}{./figure/inv_cologne_109/}{0000}{0109}
        \hfill
        \animategraphics[autoplay,loop,width=0.495\linewidth]{21}{./figure/fwd_cologne_109/}{0000}{0109}
        \caption{Reversed warping~\cite{wang2018occlusion} and forward projection.}
        \label{teaser_b}
    \end{subfigure}
    \vspace{-3mm}
    \caption{Different rendering techniques on dynamic objects. Inverse projection and reversed inverse warping cause significant appearance distortions and ghosting effects, while our forward projection technique preserves object appearance. This is a \emph{video figure}, best viewed in \emph{Adobe Reader}.}
    \label{teaser}
	\vspace{-2mm}
\end{figure}

\vspace{-0mm}
Recent advances in deep neural networks (DNNs) have led to a surge of interest in depth prediction using monocular images~\cite{eigen2014depth,garg2016unsupervised} and stereo images~\cite{mayer2016large,chang2018pyramid}, as well as in optical flow estimation~\cite{dosovitskiy2015flownet,sun2018pwc,lv2018learning}.
These supervised methods require a large amount and broad variety of training data with ground-truth labels.
Studies have shown significant progress in unsupervised learning of depth and ego-motion from unlabeled image sequences~\cite{zhou2017unsupervised,godard2017unsupervised,wang2018learning,mahjourian2018unsupervised,ranjan2019collaboration}.
The joint optimization framework uses a network for predicting single-view depth and pose, and exploits view synthesis of images in the sequence as the supervisory signal.
However, these works ignore or mask out regions of moving objects for pose and depth inference.

\vspace{-0mm}
In this work, rather than considering these moving objects as nuisances 
under the \textbf{\textit{assumption of static structure}}, 
we utilize them as important clues for estimating 3D object motions.
This problem can be formulated as factorization of object and camera motion.
Factorizing object motion in monocular sequences is a challenging problem, especially in complex urban environments that contain numerous dynamic objects.

\vspace{-0mm}
To address this problem, we propose a novel framework that explicitly models 3D motions of dynamic objects and ego-motion together with scene depth in a monocular camera setting.
Our unsupervised method relies solely on monocular video for training (without any geometric ground-truth labels) and imposes a unified photometric and geometric consistency loss on synthesized frames from one time step to the next in a sequence.
Given two consecutive frames in a video, the proposed neural network produces depth, 6-DoF motion of each moving object, and the ego-motion between adjacent frames. 
In this process, we leverage the instance mask of each dynamic object, obtained from off-the-shelf instance segmentation and optical flow modules.

\vspace{-0mm}
Our main contributions are the following:

\vspace{-0mm}
\noindent\textbf{Neural forward projection}
Differentiable depth-based rendering (which we call inverse warping) was introduced in~\cite{zhou2017unsupervised}, where the target view $I_t$ is reconstructed by sampling pixels from a source view $I_s$ based on the target depth map $D_t$ and the relative pose $T_{t\rightarrow s}$.
The warping procedure is effective in static scene areas, but the regions of moving objects cause warping artifacts because the 3D structure of the source image $I_s$ may become distorted after warping based on the target image's depth $D_t$~\cite{casser2019depth} as shown in \figref{teaser_a}.
To build a geometrically plausible formulation, we introduce forward warping (or projection) which maps the source image $I_s$ to the target viewpoint based on the source depth $D_s$ and the relative pose $T_{s\rightarrow t}$.
\footnote{This is different from the reversed optical flow leveraged in~\cite{liu2019selflow,wang2019unos,luo2019every}.
Since flow-based warping techniques do not consider geometric structure, serious distortions will appear where multiple source pixels are warped to the same target locations, \eg, object boundaries, as shown in \figref{teaser_b}.
Our \emph{forward} and \emph{inverse warping} are not about temporal order, but rather which coordinate frame from which to conduct the geometric transformation when warping from the reference to the target view.
Hereafter, we express \emph{forward projection} as \emph{forward warping} for consistency with \emph{inverse warping}.}
There is a well-known remaining issue with forward warping that the output image may have holes.
Thus, we propose the differentiable and hole-free forward warping module that works as a key component in our instance-wise depth and motion learning from monocular videos.

\vspace{-0mm}
\noindent\textbf{Instance-aware photometric and geometric consistency}
Existing works~\cite{cao2019learning,lee2019learning,liu2020flow2stereo} have successfully estimated independent object motion with stereo cameras.
Approaches based on stereo video can explicitly separate static and dynamic motion by using stereo offset and temporal information.
On the other hand, estimation from monocular video captured in the dynamic real world, where both agents and objects are moving, suffers from \emph{motion ambiguity}, as only temporal clues are available.
To address this issue, we introduce instance-aware view synthesis and unified projection consistency into the training loss.
We first decompose the image into background and object (potentially moving) regions using a predicted instance mask.
We then warp each component using the estimated single-view depth and camera poses to compute photometric consistency.
We also impose a geometric consistency loss for each instance that constrains the estimated geometry from all input frames to be consistent.


\vspace{-0mm}
\noindent\textbf{Auto-annotation of video instance segmentation}
We introduce a general-purpose auto-annotation scheme to generate a video instance segmentation dataset, which is expected to contribute to various areas of self-driving research.
The role of this method is similar to that of~\cite{yang2019video}, but we design a new framework that is tailored to driving scenarios on existing datasets~\cite{geiger2012we,cordts2016cityscapes}.
We modularize this task into instance segmentation~\cite{he2017mask,liu2018path} and optical flow~\cite{sun2018pwc} steps and combine each existing fine-tuned model to generate the tracked instance masks automatically.
we show the validity of adopting off-the-shelf instance segmentation and optical flow models without fine-tuning for our instance-wise depth and motion learning. 

\vspace{-0mm}
\noindent\textbf{State-of-the-art performance} Our self-supervised monocular depth and pose estimation is validated with a performance evaluation which shows that our jointly learned system outperforms earlier approaches.



\vspace{-0mm}
\section{Related Works}
\vspace{-0mm}

\noindent\textbf{Unsupervised depth and ego-motion learning}
Several works~\cite{zhou2017unsupervised,wang2018learning,mahjourian2018unsupervised,ranjan2019collaboration,pillai2019superdepth} have studied joint self-supervised learning of depth and ego-motion from monocular sequences with the basic concept of \emph{Structure-from-Motion (SfM)}.
Zhou~\etal~\cite{zhou2017unsupervised} introduce a unsupervised learning framework for depth and ego-motion by maximizing photometric consistency across monocular video frames during training.
Along with photo-consistency, several works~\cite{mahjourian2018unsupervised,bian2019unsupervised,chen2019self} impose geometric constraints between nearby frames with a static structure assumption.
Semantic knowledge is also used to enhance the feature representation for monocular depth estimation~\cite{chen2019towards,guizilini2020semantically}.
Recently, Guizilini~\etal~\cite{guizilini20203d} introduce a detail-preserving representation using 3D convolutions.

The aforementioned studies have a limitation on dealing with moving objects due to the rigidity assumption, which leads to performance degradation in estimating object depths.
To handle this, stereo pairs are leveraged during the training process as an auxiliary as presented by Godard~\etal~\cite{godard2017unsupervised} and Hur~\etal~\cite{hur2020self}.
With this stereo pair, every pixel correspondence between the left and right frames is described by a single camera rectification.
Please note that the monocular-based approaches are differentiated from the methodology of learning through stereo videos.

\begin{figure*}[t]
\centering
\includegraphics[width=0.99\textwidth]{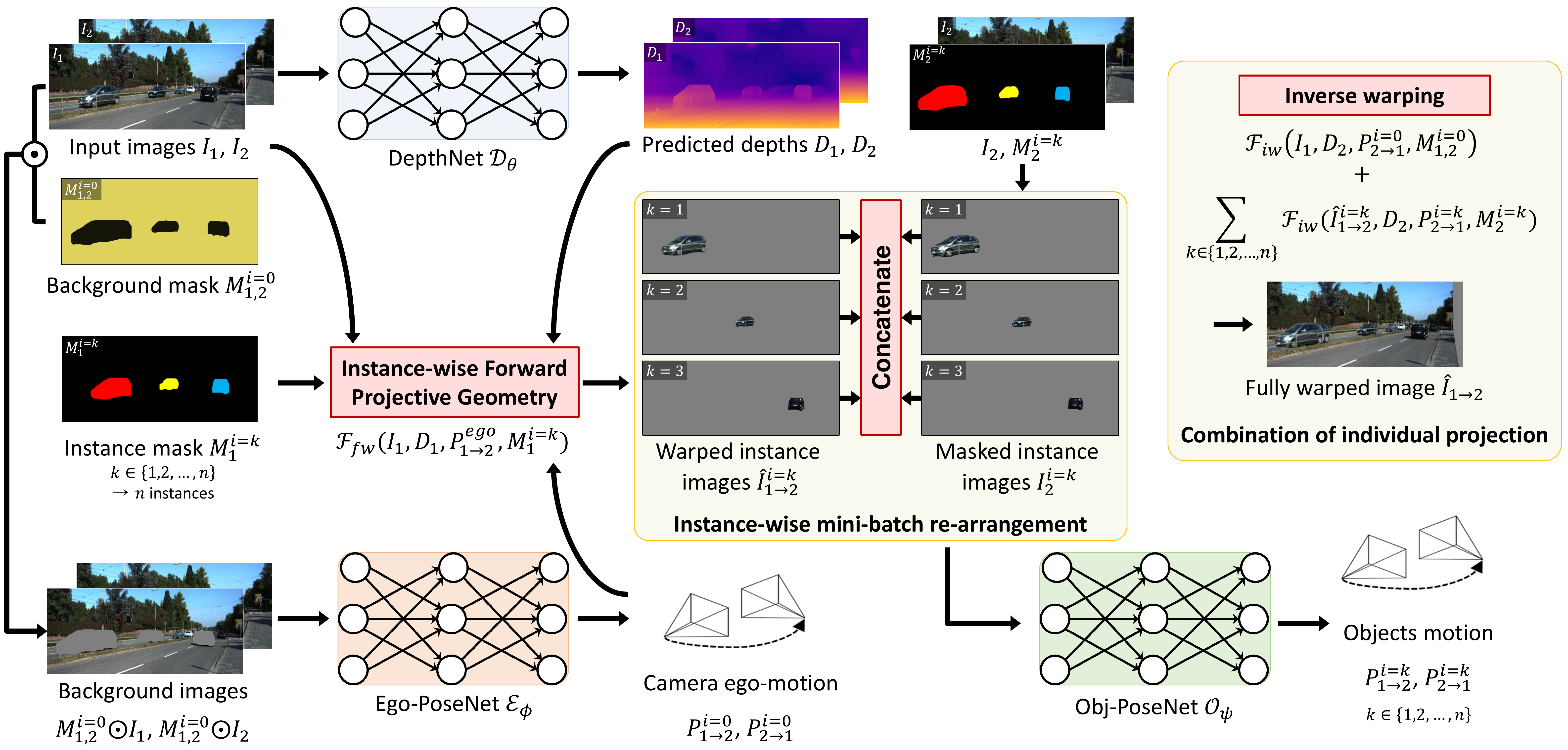}
\vspace{-2mm}
\caption{Overview of the proposed frameworks.}
\vspace{-2mm}
\label{overview}
\end{figure*}

\vspace{-0mm}
\noindent\textbf{Learning motion of moving objects}
Recently, the joint optimization of dynamic object motion along with depth and ego-motion has gained interest as a new research topic.
Cao~\etal~\cite{cao2019learning} propose a self-supervised framework with a given 2D bounding box to learn scene structure and 3D object motion from \emph{stereo} videos. The disparity from the paired images, which is \emph{deterministic}, enables computing the 3D motion vector of each instance using simple mean filtering.
Gordon~\etal~\cite{gordon2019depth} and Li~\etal~\cite{li2020unsupervised} propose a motion field network to estimate a pixel-wise transformation. It receives two consecutive rough images, which are, however, ambiguous and unclear inputs to explicitly disentangle the motion of the camera and non-rigid objects. Hence, we suggest to design the network to determine the object motion by looking at the residual signal between two images caused by pure object motion.
Casser~\etal~\cite{casser2019depth,casser2019unsupervised} and Klingner~\etal~\cite{klingner2020selfsupervised} present an unsupervised image-to-depth framework that models the motion of moving objects and cameras with given segmentation knowledge.

All the aforementioned studies use the inverse warping technique when rendering dynamic objects, which causes appearance distortion, illustrated in \figref{teaser}. Thus, we propose a \emph{geometrically correct} projection method in dynamic situations, which is a fundamental problem in 3D geometry.

\vspace{-0mm}
\section{Methodology}
\label{sec:3}
\vspace{-0mm}

We introduce an end-to-end joint training framework for instance-wise depth and motion learning from monocular videos without supervision as illustrated in \figref{overview}.
Our main contribution lies in applying the inverse and forward warping in appropriate projection situations.
In this section, we introduce the instance-wise forward projective geometry and the networks for each type of output: DepthNet, Ego-PoseNet, and Obj-PoseNet.
Further, we describe our novel loss functions and how they are designed for back-propagation in decomposing the background and moving object regions.

\vspace{-0mm}
\subsection{Method Overview}
\label{sec:3_1}
\vspace{-0mm}

\noindent\textbf{Baseline}~
Given two consecutive RGB images~$(I_1, I_2)\in \mathbb{R}^{H \times W \times 3}$, sampled from an unlabeled video, we first predict their respective depth maps~$(D_1, D_2)$ via our presented DepthNet~$\mathcal{D}_\theta: \mathbb{R}^{H \times W \times 3} \rightarrow \mathbb{R}^{H \times W}$ with trainable parameters~$\theta$.
By concatenating two sequential images as an input, our proposed Ego-PoseNet~$\mathcal{E}_\phi: \mathbb{R}^{2 \times H \times W \times 3} \rightarrow \mathbb{R}^{6}$, with trainable parameters~$\phi$, estimates the six-dimensional SE(3) relative transformation vectors~$(P_{1 \rightarrow 2}, P_{2 \rightarrow 1})$.
With the predicted depth, relative ego-motion, and a given camera intrinsic matrix~$K \in \mathbb{R}^{3 \times 3}$, we can synthesize an adjacent image in the sequence using an inverse warping operation~$\mathcal{F}_{iw}(I_i, D_j, P_{j \rightarrow i}, K) \rightarrow \hat{I}_{i \rightarrow j}$, where $\hat{I}_{i \rightarrow j}$ is the reconstructed image by warping the reference frame~$I_i$~\cite{zhou2017unsupervised,jaderberg2015spatial}.
As a supervisory signal, an image reconstruction loss, $\mathcal{L}_{rec}=||I_j - \hat{I}_{i \rightarrow j}||_1$, is imposed to optimize the parameters, $\theta$ and $\phi$.

\vspace{-0mm}
\noindent\textbf{Instance-wise learning}~
The baseline method has a limitation that it cannot handle dynamic scenes containing moving objects.
Our goal is to learn depth and ego-motion, as well as object motions, using monocular videos by constraining them with instance-wise geometric consistencies. We propose an Obj-PoseNet~$\mathcal{O}_\psi: \mathbb{R}^{2 \times H \times W \times 3} \rightarrow \mathbb{R}^{6}$ with trainable parameters~$\psi$, which is specialized to estimate individual object motions.
We annotate a novel video instance segmentation dataset to utilize it as an individual object mask while training the ego-motion and object motions.
Given two consecutive binary instance masks~$(M^{i}_1, M^{i}_2)\in \{0,1\}^{H \times W \times n}$ corresponding to $(I_1, I_2)$, $n$ instances are annotated and matched between the frames.
First, in the case of camera motion, potentially moving objects are masked out and only the background region is fed to Ego-PoseNet.
Secondly, the $n$ binary instance masks are multiplied to the input images and fed to Obj-PoseNet.
For both networks, motions of the $k^{th}$ element are represented as $P^{i=k}_{1 \rightarrow 2}$, where $k=0$ indicates camera motion from frame $I_1$ to $I_2$.
The details of the motion models will be described in the following subsections.

\begin{figure*}[t]
    \centering
    \begin{subfigure}[b]{0.33\textwidth}
        \centering
        \includegraphics[width=1\textwidth]{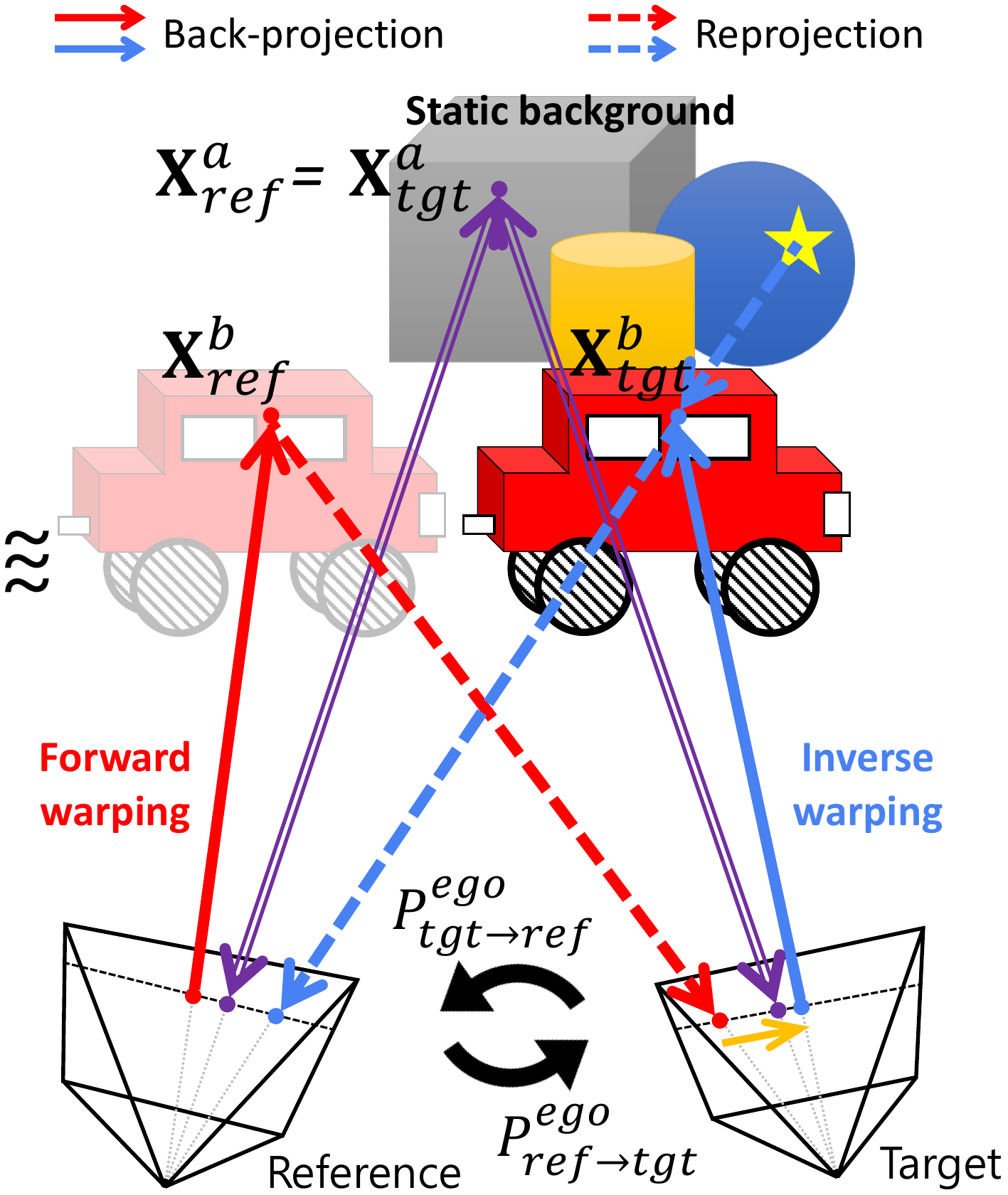}
        \caption{}
        \label{warping}
    \end{subfigure}
    \hfill
    \begin{subfigure}[b]{0.33\textwidth}
        \centering
        \includegraphics[width=1\textwidth]{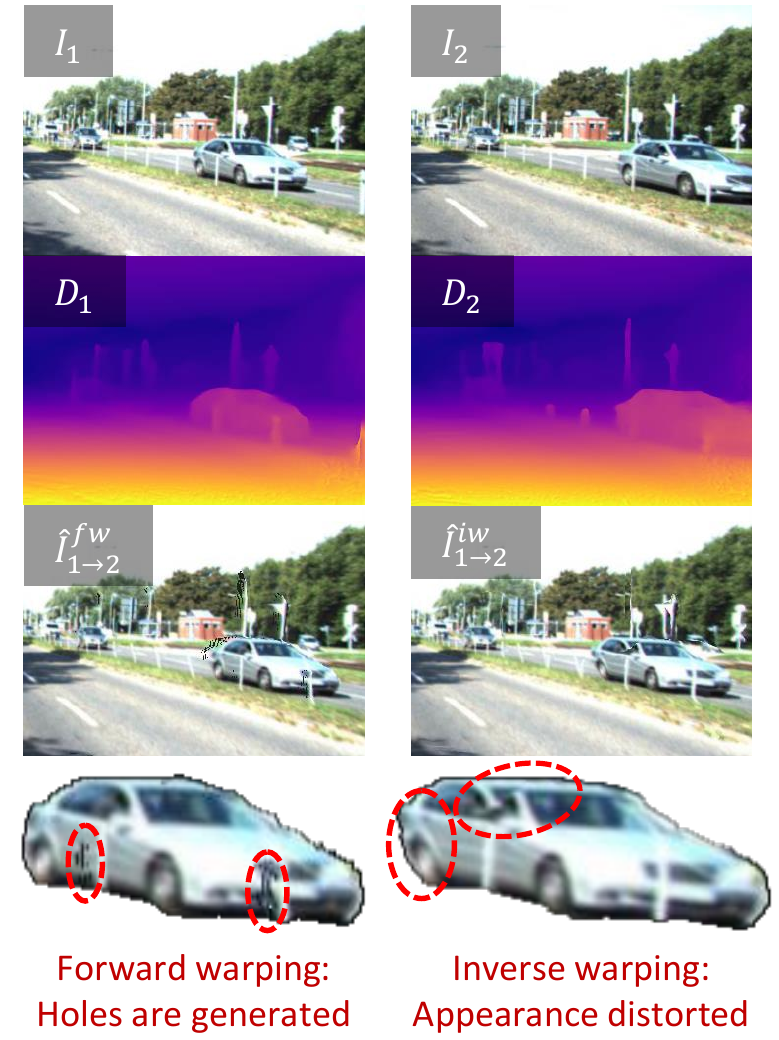}
        \caption{}
        \label{distortion_a}
    \end{subfigure}
    \hfill
    \begin{subfigure}[b]{0.33\textwidth}
        \centering
        \includegraphics[width=1\textwidth]{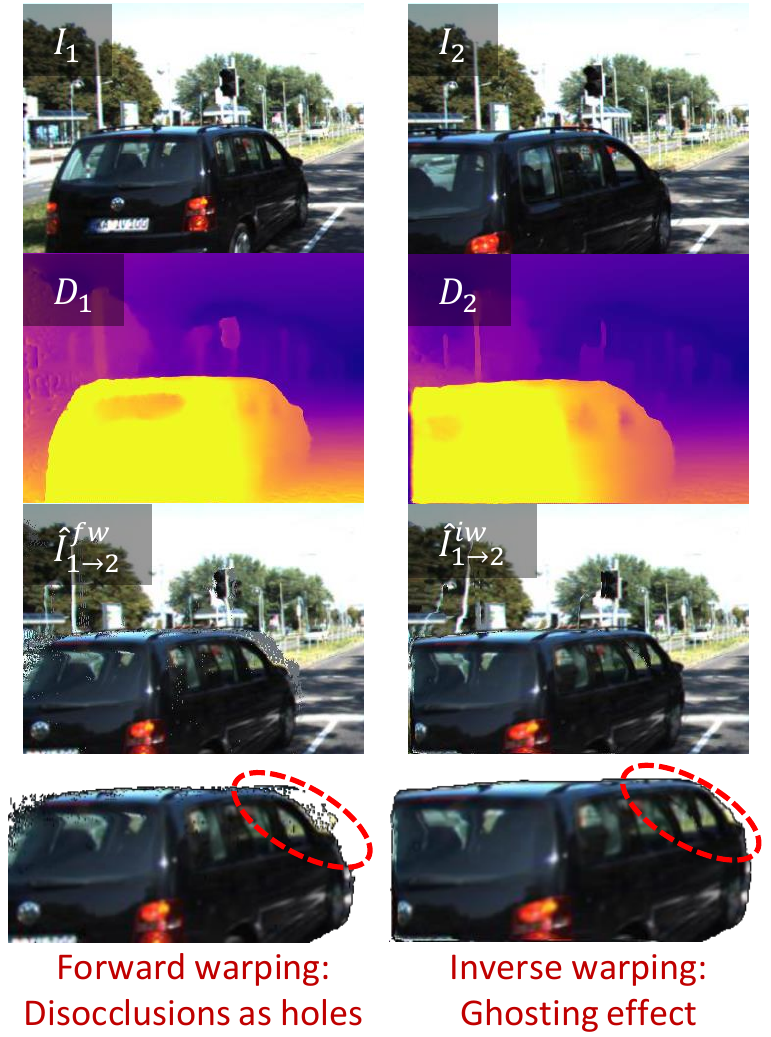}
        \caption{}
        \label{distortion_b}
    \end{subfigure}
    \vspace{-4mm}
    \caption{(a) Warping discrepancy occurs for inverse projection of moving objects. Different warping results on (b) moving and (c) close objects. $\hat{I}^{iw}$ and $\hat{I}^{fw}$ are warped only by the camera motion.}
    \label{distortion}
	\vspace{-2mm}
\end{figure*}

\vspace{-0mm}
\noindent\textbf{Training objectives}~
The previous works~\cite{mahjourian2018unsupervised,bian2019unsupervised,chen2019self,zhang2020deepptz} imposed geometric constraints between frames, but they are limited to rigid projections.
Regions containing moving objects cannot be constrained with this term and are treated as outlier regions with regard to geometric consistency.
In this paper, we propose instance-wise geometric consistency. We leverage instance masks to impose geometric consistency region-by-region.
Following instance-wise learning, our overall objective function can be defined as follows:
\begin{equation}
\Scale[0.90]
{
\begin{aligned}
\calL = \lambda_{p}\calL_{p} + \lambda_{g}\calL_{g} + \lambda_{s}\calL_{s} + \lambda_{t}\calL_{t}  + \lambda_{h}\calL_{h},
\end{aligned}
}
\label{eq_loss}
\end{equation}
where $(\calL_{p}, \calL_{g})$ are the photometric and geometric consistency losses applied on each instance region including the background, $\calL_{s}$ stands for the depth smoothness loss, and $(\calL_{t}, \calL_{h})$ are the object translation and height constraint losses.
$\{\lambda_{p}, \lambda_{g}, \lambda_{s}, \lambda_{t}, \lambda_{h}\}$ is the set of loss weights.
We train the models in both forward ($I_1 \rightarrow I_2$) and backward ($I_2 \rightarrow I_1$) directions to maximally use the self-supervisory signals.
In the following subsections, we introduce how to constrain the instance-wise consistencies.

\vspace{-0mm}
\subsection{Forward Projective Geometry}
\label{sec:3_2}
\vspace{-0mm}

A fully differentiable warping function enables learning of structure-from-motion tasks.
This operation is first proposed by \emph{spatial transformer networks} (STN)~\cite{jaderberg2015spatial}.
Previous works for learning depth and ego-motion from unlabeled videos so far follow this \emph{grid sampling} module to synthesize adjacent views.
To synthesize $\hat{I}_{1 \rightarrow 2}$ from $I_1$, the homogeneous coordinates, $p_2$, of a pixel in $I_2$ are projected to $p_1$ as follows:
\vspace{-0mm}
\begin{equation}
\Scale[0.95]
{
\begin{aligned}
p_{1} \sim K P^{i=0}_{2 \rightarrow 1} D_2(p_2) K^{-1}p_2 .
\end{aligned}
}
\label{eq_inv_warp}
\vspace{-0mm}
\end{equation}
As expressed in the equation, this operation computes $\hat{I}_{1 \rightarrow 2}$ by taking the value of the homogeneous coordinates $p_1$ from the inverse rigid projection using $P^{i=0}_{2 \rightarrow 1}$ and $D_2(p_2)$.
As a result, the coordinates $p_1$ are not valid if $p_2$ lies on an object that moves between $I_1$ and $I_2$.
Therefore, the inverse warping is not suitable for removing the effects of ego-motion in dynamic scenes.
As shown in \figref{warping}, the inverse warping causes pixel discrepancy on a moving object, since it reprojects the point ($X^b_{tgt}$) from the target geometry where the 3D point has moved.
This causes distortion of the appearance of moving objects as in \figref{distortion_a} and ghosting effects~\cite{janai2018unsupervised} on objects near to the camera as in \figref{distortion_b}.
To solve this problem, we define an intermediate frame which is transformed by camera motion with reference geometry, and mitigate the residual displacement (\textcolor{orange}{orange arrow} in \figref{warping}) by training Obj-PoseNet as a supervisory signal.
In~\tabref{tab_comp}, we describe the difference between input resources of inverse and forward warping, as well as their advantages and disadvantages.

\begin{table}[t]
	\centering
	\vspace{-0mm}
	\renewcommand{\arraystretch}{0.7}
	\caption{Comparison between inverse and forward warping.}
	\vspace{-2mm}
    \label{tab_comp}
	\resizebox{0.99\linewidth}{!}{
    	\begin{tabular}{l@{\hskip 2mm}l@{\hskip 2mm}l@{\hskip 2mm}l@{\hskip 2mm}c}
    		\toprule
             & Inverse warping & Forward warping \\ 
            \midrule
    		Inputs               & $I_{ref}$, $D_{tgt}$, $P^{ego}_{tgt\rightarrow ref}$  &  $I_{ref}$, $D_{ref}$, $P^{ego}_{ref\rightarrow tgt}$  \\ 
            Pros. 	             & Dense registration by grid sampling. & Geometry corresponds to reference.  \\ 
    		Cons.          		 & Errors induced on moving objects.      & Holes are generated.  \\
    		\bottomrule
    	\end{tabular}
	}
	\vspace{-4mm}
\end{table}


In order to synthesize the novel view (from $I_1$ to $\hat{I}_{1 \rightarrow 2}$) properly when there exist moving objects, we propose \emph{forward projective geometry}, $\mathcal{F}_{fw}(I_i, D_i, P_{i \rightarrow j}, K) \rightarrow \hat{I}_{i \rightarrow j}$ as follows:
\vspace{-0mm}
\begin{equation}
\Scale[0.95]
{
\begin{aligned}
p_{2} \sim K P^{i=0}_{1 \rightarrow 2} D^{\uparrow}_1(p_1) (K^{\uparrow})^{-1} p_1 .
\end{aligned}
}
\label{eq_fwd_warp}
\end{equation}
Unlike inverse projection in \eqnref{eq_inv_warp}, this warping process cannot be sampled by the STN since the projection is in the forward direction (inverse of \emph{grid sampling}).
In order to make this operation differentiable, we first use sparse tensor coding to index the homogeneous coordinates $p_2$ of a pixel in $I_2$.
Invalid coordinates (exiting the view where $p_2 \notin \{(x,y)| 0 \leq x<W, 0 \leq y<H\}$) of the sparse tensor are masked out.
We then convert this sparse tensor to be dense by taking the nearest neighbor value of the source pixel.
However, this process has a limitation that there exist irregular holes due to the sparse tensor coding.
Since we need to feed those forward projected images into the neural networks in the next step, the size of the holes should be minimized.
To fill these holes as much as possible, we pre-upsample the depth map~$D^{\uparrow}_1(p_1)$ of the reference frame.
If the depth map is upsampled by a factor of $\alpha$, the camera intrinsic matrix is also upsampled as follows:
\vspace{-0mm}
\begin{equation}
\Scale[0.93]
{
\begin{aligned}
K^{\uparrow} = \begin{bmatrix}
\alpha f_x & 0          & \alpha W \\
0          & \alpha f_y & \alpha H \\
0          & 0          & 1
\end{bmatrix},
\end{aligned}
}
\label{eq_intrinsics}
\end{equation}
where $(f_x, f_y)$ are the focal lengths along the $x$- and $y$-axis.
In the following subsection, we describe the steps of how to synthesize novel views with inverse and forward projection in each instance region.



\vspace{-0mm}
\subsection{Instance-Aware View Synthesis and Projection Consistency}
\label{sec:3_3}
\vspace{-0mm}

\noindent\textbf{Instance-wise projection}~
Each step of the instance-wise view synthesis is depicted in \figref{overview}.
To synthesize a novel view in an instance-wise manner, we first decompose the image region into background and object (potentially moving) regions.
With given instance masks $(M^{i}_1, M^{i}_2)$, the background mask along frames $(I_1, I_2)$ is generated as
\vspace{-0mm}
\begin{equation}
\Scale[0.84]
{
\begin{aligned}
M^{i=0}_{1,2} = (1 - \cup_{k \in \{1,2,...,n\}} M^{i=k}_1) \cap (1 - \cup_{k \in \{1,2,...,n\}} M^{i=k}_2) .
\end{aligned}
}
\label{eq_bg_mask}
\end{equation}
The background mask is pixel-wise multiplied ($\odot$) to the input frames $(I_1, I_2)$, and then concatenated along the channel axis, which is an input to Ego-PoseNet. The camera motion is computed as
\vspace{-1mm}
\begin{equation}
\Scale[0.90]
{
\begin{aligned}
P^{i=0}_{1 \rightarrow 2}, P^{i=0}_{2 \rightarrow 1} =
    \mathcal{E}_\phi( M^{i=0}_{1,2} \odot I_1, M^{i=0}_{1,2} \odot I_2).
\end{aligned}
}
\label{eq_ego_pose}
\vspace{-0mm}
\end{equation}
To learn the object motions, we first apply the forward warping, $\calF_{fw}(\cdot)$, to generate ego-motion-eliminated warped images and masks as follows:
\vspace{-0mm}
\begin{equation}
\Scale[0.90]
{
\begin{aligned}
\hat{I}^{fw}_{1 \rightarrow 2} = \calF_{fw}(I_1, D^{\uparrow}_1, P^{i=0}_{1 \rightarrow 2}, K^{\uparrow}) , \\
\hat{M}^{fw}_{1 \rightarrow 2} = \calF_{fw}(M_1, D^{\uparrow}_1, P^{i=0}_{1 \rightarrow 2}, K^{\uparrow}) .
\end{aligned}
}
\label{eq_fwd_warp}
\end{equation}
Now we can generate forward-projected instance images as $\hat{I}^{fw, i=k}_{1 \rightarrow 2} = \hat{M}^{fw,i=k}_{1 \rightarrow 2} \odot \hat{I}^{fw}_{1 \rightarrow 2} $ and $\hat{I}^{fw, i=k}_{2 \rightarrow 1} = \hat{M}^{fw,i=k}_{2 \rightarrow 1} \odot \hat{I}^{fw}_{2 \rightarrow 1} $.
For every object instance in the image, Obj-PoseNet predicts the $k^{th}$ object motion as
\vspace{-0mm}
\begin{equation}
\Scale[0.90]
{
\begin{aligned}
P^{i=k}_{1 \rightarrow 2}, P^{i=k}_{2 \rightarrow 1} =
    \mathcal{O}_\psi(\hat{I}^{fw, i=k}_{1 \rightarrow 2}, M^{i=k}_2 \odot I_2) ,
\end{aligned}
}
\label{eq_obj_pose}
\end{equation}
where both object motions are composed of six-dimensional SE(3) translation and rotation vectors.
We merge all instance regions to synthesize the novel view.
In this step, we utilize inverse warping, $\calF_{iw}(\cdot)$.
First, the background region is reconstructed as
\vspace{-1mm}
\begin{equation}
\Scale[0.90]
{
\begin{aligned}
\hat{I}^{iw, i=0}_{1 \rightarrow 2} = M^{i=0}_{1,2} \odot \calF_{iw}(I_1, D_2, P^{i=0}_{2 \rightarrow 1}, K) ,
\end{aligned}
}
\label{eq_bg_full}
\end{equation}
where the gradients are propagated with respect to $\theta$ and $\phi$.
Second, the inverse-warped $k^{th}$ instance is represented as
\vspace{-1mm}
\begin{equation}
\Scale[0.90]
{
\begin{aligned}
\hat{I}^{fw \rightarrow iw, i=k}_{1 \rightarrow 2} =
     \calF_{iw}(\hat{I}^{fw,i=k}_{1 \rightarrow 2}, D_2, P^{i=k}_{2 \rightarrow 1}, K) ,
\end{aligned}
}
\label{eq_obj_full}
\end{equation}
where the gradients are propagated with respect to $\theta$ and $\psi$.
Finally, our instance-wise fully reconstructed novel view is formulated as
\vspace{-1mm}
\begin{equation}
\Scale[0.90]
{
\begin{aligned}
\hat{I}_{1 \rightarrow 2} =
     \hat{I}^{iw, i=0}_{1 \rightarrow 2} + \sum\limits_{k \in \{1,2,...,n\}} \hat{I}^{fw \rightarrow iw, i=k}_{1 \rightarrow 2} .
\end{aligned}
}
\label{eq_full}
\end{equation}


\vspace{-0mm}
\noindent\textbf{Instance mask propagation}~
Through the process of forward and inverse warping, the instance mask is also propagated to represent information on instance position and pixel validity.
In the case of the $k^{th}$ instance mask $M^{i=k}_1$, the forward and inverse warped mask is expressed as follows:
\vspace{-0mm}
\begin{equation}
\Scale[0.90]
{
\begin{aligned}
\hat{M}^{fw \rightarrow iw, i=k}_{1 \rightarrow 2} =
     \calF_{iw}(\hat{M}^{fw,i=k}_{1 \rightarrow 2}, D_2, P^{i=k}_{2 \rightarrow 1}, K) .
\end{aligned}
}
\label{eq_inst_mask}
\end{equation}
Note that the forward warped mask $\hat{M}^{fw,i=k}_{1 \rightarrow 2}$ has holes due to the sparse tensor coding.
To keep the binary format and avoid interpolation near the holes while inverse warping, we round up the fractional values after each warping operation.
The final valid instance mask is expressed as follows:
\vspace{-1mm}
\begin{equation}
\Scale[0.90]
{
\begin{aligned}
\hat{M}_{1 \rightarrow 2} =
     M^{i=0}_{1,2} + \sum\limits_{k \in \{1,2,...,n\}} \hat{M}^{fw \rightarrow iw, i=k}_{1 \rightarrow 2} .
\end{aligned}
}
\label{eq_mask_sum}
\end{equation}


\vspace{-0mm}
\noindent\textbf{Instance-wise geometric consistency}~
We impose the geometric consistency loss for each region of an instance.
With the predicted depth map and warped instance mask, $D_1$ can be spatially aligned to the frame $D_2$ by forward and inverse warping, represented as $M^{i=0}_{1,2} \odot \hat{D}^{iw, i=0}_{1 \rightarrow 2}$ and $\hat{M}^{fw \rightarrow iw, i=k}_{1 \rightarrow 2} \odot \hat{D}^{fw \rightarrow iw, i=k}_{1 \rightarrow 2}$ respectively for background and instance regions.
In addition, $D_2$ can be scale-consistently transformed to the frame $D_1$, represented as $M^{i=0}_{1,2} \odot D^{sc, i=0}_{2 \rightarrow 1}$ and $\hat{M}^{fw \rightarrow iw, i=k}_{1 \rightarrow 2} \odot D^{sc, i=k}_{2 \rightarrow 1}$ respectively for background and instance regions.
Based on this instance-wise operation, we compute the unified depth inconsistency map as:
\vspace{-0mm}
\begin{equation}
\Scale[0.90]
{
\begin{aligned}
D^{diff, i=0}_{1 \rightarrow 2} = M^{i=0}_{1,2} \odot
     \frac{ |\hat{D}^{iw, i=0}_{1 \rightarrow 2} - D^{sc, i=0}_{2 \rightarrow 1}| }
          {  \hat{D}^{iw, i=0}_{1 \rightarrow 2} + D^{sc, i=0}_{2 \rightarrow 1}  } , \\
D^{diff, i=k}_{1 \rightarrow 2} = \hat{M}^{fw \rightarrow iw, i=k}_{1 \rightarrow 2} \odot
     \frac{ |\hat{D}^{fw \rightarrow iw, i=k}_{1 \rightarrow 2} - D^{sc, i=k}_{2 \rightarrow 1}| }
          {  \hat{D}^{fw \rightarrow iw, i=k}_{1 \rightarrow 2} + D^{sc, i=k}_{2 \rightarrow 1}  } .
\end{aligned}
}
\label{eq_gc_bg}
\end{equation}
Note that the above depth inconsistency maps are spatially aligned to the frame $D_2$.
Therefore, we can integrate the depth inconsistency maps from the background and instance regions as follows:
\vspace{-1mm}
\begin{equation}
\Scale[0.90]
{
\begin{aligned}
D^{diff}_{1 \rightarrow 2} = D^{diff, i=0}_{1 \rightarrow 2} + \sum\limits_{k \in \{1,2,...,n\}} D^{diff, i=k}_{1 \rightarrow 2} .
\end{aligned}
}
\label{eq_gc}
\end{equation}

\vspace{-0mm}
\noindent\textbf{Training loss}~
In order to handle occluded, view-exiting, and invalid instance regions, we leverage \eqnref{eq_mask_sum} and \eqnref{eq_gc}.
We generate a weight mask as $1-D^{diff}_{1 \rightarrow 2}$ and this is multiplied to the valid instance mask $\hat{M}_{1 \rightarrow 2}$.
Finally, our weighted valid mask is formulated as:
\vspace{-1mm}
\begin{equation}
\Scale[0.90]
{
\begin{aligned}
V_{1 \rightarrow 2} = (1-D^{diff}_{1 \rightarrow 2}) \odot \hat{M}_{1 \rightarrow 2} .
\end{aligned}
}
\label{eq_valid}
\end{equation}
The photometric consistency loss $\calL_p$ is expressed as follows:
\vspace{-2mm}
\begin{equation}
\Scale[0.86]
{
\begin{aligned}
\calL_p =
\sum\limits_{{\rm{x}} \in X} V_{1 \rightarrow 2}({\rm{x}}) \cdot  \left\{ (1-\gamma) \cdot \left | { {I_2({\rm{x}}) - {\hat{I}_{1 \rightarrow 2}}({\rm{x}})} } \right |_{1}  \right. \\
\left.
+ \gamma \left( 1 - SSIM(I_2({\rm{x}}), {{\hat I}_{1 \rightarrow 2}}({\rm{x}})) \right)  \right\} ,
\end{aligned}
}
\label{eq_loss_recon}
\end{equation}
where $\rm{x}$ is the location of each pixel, $SSIM(\cdot)$ is the structural similarity index~\cite{wang2004image}, and $\gamma$ is set to 0.85 based on cross-validation.
The geometric consistency loss $\calL_g$ is expressed as follows:
\vspace{-1mm}
\begin{equation}
\Scale[0.90]
{
\begin{aligned}
\calL_g = \sum\limits_{{\rm{x}} \in X} \hat{M}_{1 \rightarrow 2}({\rm{x}}) \cdot D^{diff}_{1 \rightarrow 2}({\rm{x}}).
\end{aligned}
}
\label{eq_loss_gc}
\end{equation}

\vspace{-0mm}
To mitigate spatial fluctuation, we incorporate a smoothness term to regularize the predicted depth.
We apply the edge-aware smoothness loss proposed by Ranjan~\etal~\cite{ranjan2019collaboration}, which is described as:
\vspace{-0mm}
\begin{equation}
\Scale[0.90]
{
\begin{aligned}
\calL_s = \sum\limits_{{\rm{x}} \in X} ( \nabla D_1({\rm{x}}) \cdot e^{-\nabla I_1({\rm{x}})} )^2 .
\end{aligned}
}
\label{eq_loss_s}
\vspace{-0mm}
\end{equation}
Note that the above equations are imposed for both forward and backward directions by switching the subscripts ${}_1$ and ${}_2$.

\vspace{-0mm}
Since the dataset has a low proportion of moving objects, the learned motions of objects tend to converge to zero.
The same issue has been raised in a previous study~\cite{casser2019depth}.
To supervise the approximate amount of an object's movement, we constrain the motion of the object with a translation prior.
We compute this translation prior, $\vv{t_p}$, by subtracting the mean estimate of the object's 3D points in the forward warped frame into that of the target frame's 3D object points. This represents the mean estimated 3D vector of the object's motion.
The object translation constraint loss measures scale and cosine similarity of 3D vectors as follows:
\vspace{-1mm}
\begin{equation}
\Scale[0.81]
{
\begin{aligned}
\calL_t = \sum\limits_{k \in \{1,2,...,n\}} \left ( \left | \parallel \vv{t^{i=k}} \parallel - \parallel \vv{t^{i=k}_p} \parallel \right | _1 + loss_{\measuredangle}(\vv{t^{i=k}}, \vv{t^{i=k}_p})
\right ) ,
\end{aligned}
}
\label{eq_loss_t}
\end{equation}
where $\vv{t^{i=k}}$ and $\vv{t^{i=k}_p}$ are predicted object translation from Obj-PoseNet and the translation prior on the $k^{th}$ instance mask, and $loss_{\measuredangle}$ is a cosine similarity loss between 3D vectors.

\vspace{-0mm}
Although we have accounted for instance-wise geometric consistency, there still exists a trivial case of infinite depth for a moving object that has the same motion as the camera motion, such as for a vehicle in front.
To mitigate this issue, we adopt the object height constraint loss proposed by a previous study~\cite{casser2019depth}, which is described as:
\vspace{-1mm}
\begin{equation}
\Scale[0.90]
{
\begin{aligned}
\calL_h = \sum\limits_{k \in \{1,2,...,n\}} \frac{1}{\overline{D}} \cdot \left | D \odot M^{i=k} - \frac{ f_y \cdot p_h }{ h^{i=k} } \right | _1 ,
\end{aligned}
}
\label{eq_loss_h}
\vspace{-2mm}
\end{equation}
where $\overline{D}$ is the mean estimated depth, and ($p_h$, $h^{i=k}$) are a learnable height prior and pixel height of the $k^{th}$ instance.
Unlike the previous study, for stable training, the learning rate of $p_h$ is reduced to $0.1$ times and the gradient of $\overline{D}$ is detached.
The final loss is a weighted summation of the five loss terms, defined as \eqnref{eq_loss}.





\vspace{-0mm}
\section{Experiments}
\label{sec:4}
\vspace{-0mm}


\vspace{-0mm}
\subsection{Implementation Details}
\label{sec:4_1}
\vspace{-0mm}

\noindent\textbf{Network details}~
For DepthNet, we use DispResNet~\cite{ranjan2019collaboration} and a ResNet18-based encoder-decoder structure. The network can generate multi-scale outputs (six different scales), but the single-scale training converges faster and produces better performance as shown from SC-SfM~\cite{bian2019unsupervised}.
The structures of Ego-PoseNet and Obj-PoseNet are the same, but the weights are not shared. They consist of seven convolutional layers and regress the relative pose as three Euler angles and three translation vectors. 

\vspace{-0mm}
\noindent\textbf{Training}~
Our system is implemented in PyTorch~\cite{paszke2019pytorch}.
We train our networks using the ADAM optimizer~\cite{kingma2015adam} with $\beta_1 = 0.9$ and $\beta_2 = 0.999$ on $4\times$Nvidia RTX 2080 GPUs.
The image resolution is set to $832\times 256$ and the video data is augmented with random scaling, cropping, and horizontal flipping.
We set the mini-batch size to 4 and train the networks over 200 epochs with 1,000 randomly sampled batches in each epoch considering the representation capacity~\cite{zhang2019revisiting,zhang2020resnet}.
The initial learning rate is set to $10^{-4}$ and is decreased by half every 50 epochs.
The loss weights are set to $\lambda_p = 2.0$, $\lambda_g = 1.0$, $\lambda_s = 0.1$, $\lambda_t = 0.1$, and $\lambda_h = 0.02$.

\vspace{-0mm}
\noindent\textbf{Video instance segmentation dataset}~
We introduce an auto-annotation scheme to generate two video instance segmentation datasets, KITTI-VIS and Cityscapes-VIS, from existing driving datasets, KITTI~\cite{geiger2012we} and Cityscapes~\cite{cordts2016cityscapes}.
To this end, we adopt an off-the-shelf instance segmentation model, \eg, Mask R-CNN~\cite{he2017mask} and PANet~\cite{liu2018path}, and an optical flow model, PWC-Net~\cite{sun2018pwc}, for mask propagation.
We first compute the instance segmentation for every image frame, and calculate the Intersection over Union (IoU) scores among instances in each frame.
If the maximal IoU in the adjacent frame is above a threshold ($\tau = 0.5$), then the instance is assumed to be tracked and both masks are assigned with the same ID.
The occluded regions by the bidirectional consistency check~\cite{meister2017unflow} are excluded while computing the IoU scores.
The instance ID is ordered by the size of the reference instance, with the
maximum size among the matched instances coming first.
This size ordering is necessary, since we set the maximum number of instances with larger instances having a higher priority in the optimization. In our training, we set the maximum number of instances as three.

\begin{table}[t]
\centering
\caption{Ablation study (backbone - DispResNet) on KITTI Eigen split for both background (bg.) and object (obj.) areas.}
\vspace{-2mm}
\begin{adjustbox}{width=0.99\linewidth}
\setlength{\tabcolsep}{5pt}
\begin{tabular}{ccccccc}
    \Xhline{3\arrayrulewidth}
    \multirow{2}[3]{*}{\shortstack{Instance \\ knowledge}} & \multirow{2}[3]{*}{\shortstack{Geometric \\ consistency}} & \multicolumn{2}{c}{Object warping} & \multicolumn{3}{c}{AbsRel} \\
    \cmidrule(l{2pt}r{2pt}){3-4} \cmidrule(l{2pt}r{2pt}){5-7}
    &  & inverse & forward & all & bg. & obj.  \\
    \Xhline{2\arrayrulewidth}
    \xmark & \xmark & \xmark & \xmark & 0.156 & 0.142 & 0.396 \\
    \xmark & \cmark & \xmark & \xmark & \underbar{0.137} & \underbar{0.124} & 0.309 \\
    \cmark & \xmark & \cmark & \xmark & 0.151 & 0.138 & 0.377 \\
    \cmark & \cmark & \cmark & \xmark & 0.146 & 0.131 & 0.362 \\
    \cmark & \xmark & \xmark & \cmark & 0.143 & 0.133 & \underbar{0.285} \\
    \rowcolor{red!7}
    \cmark & \cmark & \xmark & \cmark & \textbf{0.124} & \textbf{0.119} & \textbf{0.178} \\
    \Xhline{3\arrayrulewidth}
\end{tabular}
\end{adjustbox}
\label{tab_ablation}
\vspace{-1mm}
\end{table}

\begin{table}[t]
\centering
\caption{Evaluation on KITTI 2015 scene flow training set. We evaluate the disparity compared to recent monocular-based training methods.}
\vspace{-2mm}
\begin{adjustbox}{width=0.99\linewidth}
\setlength{\tabcolsep}{5pt}
\begin{tabular}{lccccccc}
    \Xhline{3\arrayrulewidth}
    \multirow{2}[3]{*}{Method} & \multirow{2}{*}{Backbone} & \multicolumn{3}{c}{D1} & \multicolumn{3}{c}{D2} \\
    \cmidrule(l{2pt}r{2pt}){3-5} \cmidrule(l{2pt}r{2pt}){6-8}
     &  & bg. & fg. & all & bg. & fg. & all  \\
    \Xhline{2\arrayrulewidth}
    CC~\cite{ranjan2019collaboration}  & DispResNet & 35.0 & 42.7 & 36.2 & -- & -- & -- \\
    SC-SfM~\cite{bian2019unsupervised} & DispResNet & 36.0 & 46.5 & 37.5 & -- & -- & -- \\
    EPC++ (mono)~\cite{luo2019every}   & DispNet & 30.7 & 34.4 & 32.7 & \textbf{18.4} & 84.6 & 65.6 \\
    \rowcolor{red!7}
    Ours & DispResNet & \textbf{26.8} & \textbf{30.4} & \textbf{27.4} & 28.9      & \textbf{32.3} & \textbf{29.4} \\
    \Xhline{3\arrayrulewidth}
\end{tabular}
\end{adjustbox}
\label{tab_sceneflow}
\vspace{-3mm}
\end{table}

\begin{table*}[t]
    \centering
    \caption{Monocular depth estimation results on the KITTI (K) Eigen test split and Cityscapes (C) test set. Models pretrained on Cityscapes and fine-tuned on KITTI are denoted by `C+K'. Models trained with semantic knowledge are denoted by `S'. For each partition, best results are written in \textbf{boldface}.}
    \vspace{-3mm}
    \begin{adjustbox}{width=0.999\textwidth}
    \setlength{\tabcolsep}{5pt}
    \begin{tabular}{lccccccccccc}

    \Xhline{4\arrayrulewidth}
    \multirow{2}{*}{Method} & \multirow{2}{*}{Backbone} & \multirow{2}{*}{Training} & \multirow{2}{*}{Test} & \multicolumn{4}{c}{Error metric $\downarrow$} & & \multicolumn{3}{c}{Accuracy metric $\uparrow$}  \\
    \cline{5-8} \cline{10-12}
     &  &  &  & AbsRel  & SqRel & RMSE & RMSE log & & $\delta < 1.25$ & $\delta < 1.25^2$ & $\delta < 1.25^3$ \\ \Xhline{2\arrayrulewidth}

    EPC++~\cite{luo2019every}                   & DispNet & K & K & 0.141 & 1.029 & 5.350 & 0.216 &  & 0.816 & 0.941 & 0.976 \\
    CC~\cite{ranjan2019collaboration}           & DispResNet & K & K & 0.140 & 1.070 & 5.326 & 0.217 &  & 0.826 & 0.941 & 0.975 \\
    SC-SfM~\cite{bian2019unsupervised}          & DispResNet & K & K & 0.137 & 1.089 & 5.439 & 0.217 &  & 0.830 & 0.942 & 0.975 \\
    \rowcolor{red!7}
    Ours    & DispResNet & K (S) & K & \textbf{0.124} & \textbf{0.886} & \textbf{5.061} & \textbf{0.206} &  & \textbf{0.844} & \textbf{0.948} & \textbf{0.979} \\
    \Xhline{1\arrayrulewidth}
    
    GLNet~\cite{chen2019self}                   & ResNet18 & K     & K & 0.135 & 1.070 & 5.230 & 0.210 &  & 0.841 & 0.948 & 0.980 \\
    Monodepth2~\cite{godard2019digging}         & ResNet18 & K     & K & 0.132 & 1.044 & 5.142 & 0.210 &  & 0.845 & 0.948 & 0.977 \\
    Li~\etal~\cite{li2020unsupervised}          & ResNet18 & K     & K & 0.130 & 0.950 & 5.138 & 0.209 &  & 0.843 & 0.948 & 0.978 \\
    Struct2Depth~\cite{casser2019depth}         & ResNet18 & K (S) & K & 0.141 & 1.026 & 5.290 & 0.215 &  & 0.816 & 0.945 & 0.979 \\
    Gordon~\etal~\cite{gordon2019depth}         & ResNet18 & K (S) & K & 0.128 & 0.959 & 5.230 & 0.212 &  & 0.845 & 0.947 & 0.976 \\
    SGDepth~\cite{klingner2020selfsupervised}   & ResNet18 & K (S) & K & 0.113 & 0.835 & \textbf{4.693} & \textbf{0.191} &  & \textbf{0.879} & \textbf{0.961} & 0.981 \\
    \rowcolor{red!7}
    Ours    & ResNet18 & K (S) & K & \textbf{0.112} & \textbf{0.777} & 4.772 & \textbf{0.191} &  & 0.872 & 0.959 & \textbf{0.982} \\
    \Xhline{1\arrayrulewidth}

    CC~\cite{ranjan2019collaboration}   & DispResNet & C+K & K & 0.139 & 1.032 & 5.199 & 0.213 &  & 0.827 & 0.943 & 0.977 \\
    SC-SfM~\cite{bian2019unsupervised}  & DispResNet & C+K & K & 0.128 & 1.047 & 5.234 & 0.208 &  & 0.846 & {0.947} & 0.976 \\
    
    \rowcolor{red!7}
    Ours    & DispResNet & C+K (S) & K & \textbf{0.119} & \textbf{0.863} & \textbf{4.984} & \textbf{0.202} &  & \textbf{0.856} & \textbf{0.950} & \textbf{0.980} \\
    \Xhline{1\arrayrulewidth}
    
    Gordon~\etal~\cite{gordon2019depth}  & ResNet18 & C+K (S) & K & 0.124 & 0.930 & 5.120 & 0.206 &  & 0.851 & 0.950 & 0.978 \\
    \rowcolor{red!7}
    Ours    & ResNet18 & C+K (S) & K & \textbf{0.109} & \textbf{0.740} & \textbf{4.547} & \textbf{0.184} &  & \textbf{0.883} & \textbf{0.962} & \textbf{0.983} \\
    \Xhline{1\arrayrulewidth}
    
    Li~\etal~\cite{li2020unsupervised}          & ResNet18 & C     & C & 0.119 & 1.290 & 6.980 & 0.190 &  & 0.846 & 0.952 & 0.982 \\
    Struct2Depth~\cite{casser2019unsupervised}  & ResNet18 & C (S) & C & 0.145 & 1.737 & 7.280 & 0.205 &  & 0.813 & 0.942 & 0.978 \\
    Gordon~\etal~\cite{gordon2019depth}         & ResNet18 & C (S) & C & 0.127 & 1.330 & 6.960 & 0.195 &  & 0.830 & 0.947 & 0.981 \\
    \rowcolor{red!7}
    Ours    & ResNet18 & C (S) & C & \textbf{0.111} & \textbf{1.158} & \textbf{6.437} & \textbf{0.182} &  & \textbf{0.868} & \textbf{0.961} & \textbf{0.983} \\
    \Xhline{4\arrayrulewidth}

    \end{tabular}
    \end{adjustbox}
    \label{tab_kitti}
    \vspace{-3mm}
\end{table*}

\vspace{-0mm}
\subsection{Ablation Study}
\vspace{-0mm}

We conduct an ablation study to validate the effect of our forward projective geometry and instance-wise geometric consistency term on monocular depth estimation.
The ablation is performed with the KITTI Eigen split~\cite{eigen2014depth}.
The models are validated with the AbsRel metric by separating the background and object areas, which are masked by our annotation.
As described in \tabref{tab_ablation}, we first evaluate SC-SfM~\cite{bian2019unsupervised} as a baseline, which is not trained with instance knowledge (the $1^{st}$ and $2^{nd}$ models).
Since there are no instance masks, DepthNet is trained by inverse warping the whole image.
The results show that the geometric consistency term over the whole image plane boosts the performance of depth estimation.
With the given instance masks, we try both inverse and forward warping on the object areas. 
The inverse warping on the objects slightly improves the depth estimation; however, we observe that Obj-PoseNet does not converge (the $3^{rd}$ and $4^{th}$ models). 
Rather, the performance is degraded when using the instance-wise geometric consistency term with inverse warping on the objects (comparing the $2^{nd}$ and $4^{th}$ models).
We conjecture that the uncertainty in learning the depth of the object area degrades the performance on the background depth around which the object is moving.
However, the forward warping on the objects improves the depth estimation on both background and object areas (the $5^{th}$ and $6^{th}$ models).
This shows that a well-optimized Obj-PoseNet helps to boost the performance of DepthNet and they complement each other.
We note that the background is still inverse warped to synthesize the target view and the significant performance improvement comes from the instance-wise geometric loss incorporated with forward projection while warping the object areas.


\begin{table}[t]
\centering
\caption{Absolute trajectory error (ATE) on KITTI visual odometry.}
\vspace{-2mm}
\begin{adjustbox}{width=0.99\linewidth}
\setlength{\tabcolsep}{5pt}
\begin{tabular}{lcc}
    \Xhline{3\arrayrulewidth}
    Method & Seq. 09 & Seq. 10 \\
    \Xhline{1\arrayrulewidth}
    SfM-Learner~\cite{zhou2017unsupervised}     & $0.021\pm0.017$ & $0.020\pm0.015$ \\
    GeoNet~\cite{yin2018geonet}                 & $0.012\pm0.007$ & $0.012\pm0.009$ \\
    CC~\cite{ranjan2019collaboration}           & $0.012\pm0.007$ & $0.012\pm0.008$ \\
    Struct2Depth~\cite{casser2019depth}         & $0.011\pm0.006$ & $0.011\pm0.010$ \\
    GLNet~\cite{chen2019self}                   & $0.011\pm0.006$ & $0.011\pm0.009$ \\
    SGDepth~\cite{klingner2020selfsupervised}   & $0.017\pm0.009$ & $0.014\pm0.010$ \\
    \Xhline{1\arrayrulewidth}
    Ours (w/o inst.)                            & $0.012\pm0.008$ & $0.011\pm0.010$ \\
    Ours (w/ inst.)                             & $\mathbf{0.010\pm0.013}$ & $\mathbf{0.011\pm0.008}$ \\
    \Xhline{3\arrayrulewidth}
\end{tabular}
\end{adjustbox}
\label{tab_odom}
\vspace{-1mm}
\end{table}

\begin{table}[t]
\centering
\caption{Relative translation $t_{err}$ ($\%$) and rotation $r_{err}$ ($\sfrac{^{\circ}}{100m}$) errors on KITTI visual odometry.}
\vspace{-2mm}
\begin{adjustbox}{width=0.69\linewidth}
\setlength{\tabcolsep}{5pt}
\begin{tabular}{lcccc}
    \Xhline{3\arrayrulewidth}
    \multirow{2}[3]{*}{Method} & \multicolumn{2}{c}{Seq. 09} & \multicolumn{2}{c}{Seq. 10} \\
    \cmidrule(l{2pt}r{2pt}){2-3} \cmidrule(l{2pt}r{2pt}){4-5}
    & $t_{err}$ & $r_{err}$ & $t_{err}$ & $r_{err}$  \\
    \Xhline{2\arrayrulewidth}
    GeoNet~\cite{yin2018geonet} & 39.4 & 14.3 & 29.0 & 8.6 \\
    SC-SfM~\cite{bian2019unsupervised} & 11.2 & 3.4 & 10.1 & 5.0 \\
    \Xhline{1\arrayrulewidth}
    Ours (w/o inst.) & 10.2 & 5.2 & 10.1 & 4.8 \\
    Ours (w/ inst.) & \textbf{8.6} & \textbf{2.9} & \textbf{9.2} & \textbf{4.5} \\
    \Xhline{3\arrayrulewidth}
\end{tabular}
\end{adjustbox}
\label{tab_odom2}
\vspace{-3mm}
\end{table}

\vspace{0mm}
\subsection{Monocular Depth Estimation}
\vspace{-0mm}

\noindent\textbf{Test setup}~
First, we show the disparity results on the KITTI 2015 scene flow training set. Our models are trained with non-overlapped KITTI raw images. We follow the standard metrics (D1, D2: percentage of erroneous pixels over all pixels). 
Since monocular training has a scale issue, we assume that the scale for disparity is given, which is the same experimental setup in EPC++~\cite{luo2019every}.

Second, we train and test our models with the Eigen split~\cite{eigen2014depth} of the KITTI dataset, and the Cityscapes dataset following the method from Struct2Depth~\cite{casser2019depth}.
We compare the performance of the proposed method with recent state-of-the-art works~\cite{chen2019self,casser2019depth,bian2019unsupervised,ranjan2019collaboration,godard2019digging,gordon2019depth,klingner2020selfsupervised,li2020unsupervised} for unsupervised single-view depth estimation.


\vspace{-0mm}
\noindent\textbf{Results analysis}~
\tabref{tab_sceneflow} shows the results on KITTI 2015 scene flow. The foreground (fg.) results show the superiority on handling dynamic regions. 
\tabref{tab_kitti} shows the KITTI Eigen split and Cityscapes test results, where ours achieves state-of-the-art performance in the single-view depth prediction task with unsupervised monocular training. 
The advantage is evident from using instance masks and constraining the instance-wise photometric and geometric consistencies.
Note that we do not need instance masks for DepthNet in testing.


\vspace{-0mm}
\subsection{Visual Odometry}
\vspace{-0mm}

\noindent\textbf{Test setup}~
We evaluate the performance of our Ego-PoseNet on the KITTI visual odometry dataset.
Following the evaluation setup of SfM-Learner~\cite{zhou2017unsupervised}, we use sequences 00-08 for training, and sequences 09 and 10 for tests.
In our case, since the potentially moving object masks are fed together with the image sequences while training Ego-PoseNet, we test the performance of visual odometry under two conditions: with and without instance masks.

\vspace{-0mm}
\noindent\textbf{Results analysis}~
We measure the absolute trajectory error (ATE) in \tabref{tab_odom} and relative errors ($t_{err}$, $r_{err}$) in \tabref{tab_odom2}, which show state-of-the-art performance.
Although we do not use the instance mask, the result of sequence 10 produces favorable performance.
This is because the scene does not have many potentially moving objects, \eg, vehicles and pedestrians, so the result is not affected much by using or not using instance masks.

\vspace{-0mm}
\section{Conclusion}
\vspace{-0mm}
In this work, we propose a unified framework that predicts monocular depth, ego-motion, and 6-DoF motion of multiple dynamic objects by training on monocular videos.
Leveraging video instance segmentation, we design an end-to-end joint training pipeline.
There are three main contributions of our work: (1) a neural forward projection module, (2) a unified instance-aware photometric and geometric consistency loss, and (3) an auto-annotation scheme for video instance segmentation.
We show that our method outperforms the existing unsupervised methods that estimate monocular depth.
We also show that each proposed module plays a role in improving the performance of our framework.



\subsection{Acknowledgements}
This research was supported by the Shared Sensing for Cooperative Cars Project funded by Bosch (China) Investment Ltd.
This work was supported by the National Research Foundation of Korea (NRF) grant funded by the Korea government (MSIT) (No. 2020R1C1C1013210), and the DGIST R\&D Program of the Ministry of Science and ICT (20-CoE-IT-01).

\bibliography{egbib}

\end{document}